\documentclass[a4paper,fleqn]{cas-sc}

\usepackage[authoryear,longnamesfirst]{natbib}

\usepackage{tikz-dependency}
\usepackage{tikz}
\usepackage{tikz-qtree}
\usepackage{latexsym}
\usepackage{amssymb}
\usepackage{amsmath}
\usepackage{arydshln}

\defcitealias{L2RPointer}{\scshape FG\&GR19}
\defcitealias{DiscoPointer}{\scshape FG\&GR20}

\def\tsc#1{\csdef{#1}{\textsc{\lowercase{#1}}\xspace}}
\tsc{WGM}
\tsc{QE}
\defcitealias{fernandez-gonzalez-gomez-rodriguez-2019-left}{Fdez-G \& G\'omez-R (2019)}
\defcitealias{multipointer}{Fdez-G \& G\'omez-R (2020)}

\newcommand{\dani}[1]{\textcolor{black}{#1}}
\newcommand{\new}[1]{\textcolor{black}{#1}}

\newcommand{\added}[1]{\textcolor{black}{#1}}

\begin{document}
\let\WriteBookmarks\relax
\def\floatpagepagefraction{1}
\def\textpagefraction{.001}

% Short title
\shorttitle{Multitask Pointer Network for Multi-Representational Parsing}    

% Short author
\shortauthors{D. Fern\'andez-Gonz\'alez, C. G\'omez-Rodr\'iguez.}  

% Main title of the paper
\title [mode = title]{Multitask Pointer Network for Multi-Representational Parsing}  

% Title footnote mark
% eg: \tnotemark[1]
%\tnotemark[<tnote number>] 

% Title footnote 1.
% eg: \tnotetext[1]{Title footnote text}
%\tnotetext[<tnote number>]{<tnote text>} 

% First author
%
% Options: Use if required
% eg: \author[1,3]{Author Name}[type=editor,
%       style=chinese,
%       auid=000,
%       bioid=1,
%       prefix=Sir,
%       orcid=0000-0000-0000-0000,
%       facebook=<facebook id>,
%       twitter=<twitter id>,
%       linkedin=<linkedin id>,
%       gplus=<gplus id>]

\author[1]{Daniel Fern\'{a}ndez-Gonz\'{a}lez}[orcid=0000-0002-6733-2371]

% Corresponding author indication
\cormark[1]

% Footnote of the first author
%\fnmark[<footnote mark no>]

% Email id of the first author
\ead{d.fgonzalez@udc.es}

% URL of the first author
\ead[url]{https://danifg.github.io}

% Credit authorship
% eg: \credit{Conceptualization of this study, Methodology, Software}
\credit{Conceptualization, methodology, software, validation, formal analysis, investigation, data curation, writing - original draft, writing - review \& editing, visualization}
%\credit{<Credit authorship details>}

% Address/affiliation
\affiliation[1]{organization={Universidade da Coru\~{n}a, CITIC, FASTPARSE Lab, LyS Group, Depto. de Ciencias de la Computaci\'{o}n y Tecnolog\'{i}as de la Informaci\'{o}n},
            addressline={Campus de Elvi\~{n}a, s/n }, 
            city={A Coru\~{n}a},
%          citysep={}, % Uncomment if no comma needed between city and postcode
            postcode={15071}, 
            %state={},
            country={Spain}}

\author[1]{Carlos G\'{o}mez-Rodr\'{i}guez}[orcid=0000-0003-0752-8812]

% Footnote of the second author
%\fnmark[2]

% Email id of the second author
\ead{carlos.gomez@udc.es}

% URL of the second author
\ead[url]{http://www.grupolys.org/~cgomezr/}

% Credit authorship
\credit{Conceptualization, validation, formal analysis, writing - review \& editing, supervision, project administration, funding acquisition}

% Address/affiliation
% \affiliation[<aff no>]{organization={},
%             addressline={}, 
%             city={},
% %          citysep={}, % Uncomment if no comma needed between city and postcode
%             postcode={}, 
%             state={},
%             country={}}

% Corresponding author text
\cortext[1]{Corresponding author}

% Footnote text
%\fntext[1]{}

% For a title note without a number/mark
%\nonumnote{}

% Here goes the abstract
\begin{abstract}
Dependency and constituent trees are widely used by many artificial intelligence applications for representing the syntactic structure of human languages. Typically, these structures are separately produced by either dependency or constituent parsers. In this article,
we propose a transition-based approach that, by training a single model, can efficiently parse any input sentence with both constituent and dependency trees,
supporting both continuous/projective and discontinuous/non-projective syntactic structures. To that end, we develop a Pointer Network architecture with two separate task-specific decoders and a common encoder, and follow a multitask learning strategy to jointly train them. The resulting quadratic system, not only becomes the first parser that can jointly produce both unrestricted constituent 
and dependency  
trees from a single model, but also proves that both syntactic formalisms can benefit from each other during training, achieving state-of-the-art accuracies in several widely-used benchmarks such as the continuous English and Chinese Penn Treebanks, as well as the discontinuous German NEGRA and TIGER datasets. 
\end{abstract}

% Use if graphical abstract is present
%\begin{graphicalabstract}
%\includegraphics{}
%\end{graphicalabstract}

% Keywords
% Each keyword is seperated by \sep

\maketitle

% Main text
\section{Introduction}
\new{Numerous artificial intelligence systems that demand natural language processing of texts and speech} are currently using syntactic formalisms for representing the grammatical structure of sentences. Among them, we can find those that recently use it for machine translation \citep{zhang-etal-2019-syntax-enhanced,YANG2020105042,ZHANG2021103427}, relation and event extraction \citep{Nguyen2019}, opinion mining \citep{zhang-etal-2020-syntax,xia-etal-2021-unified}, question answering \citep{CAOPMID:31562071,xu2021syntaxenhanced}, sentence classification \citep{ZHANG2021103427}, sentiment classification \citep{bai-etal-2021-syntax}, summarization \citep{balachandran-etal-2021-structsum} or semantic role labeling and named entity recognition \citep{sachan-etal-2021-syntax}. 
To that end, 
two widely-known formalisms are commonly used: \textit{constituent} and \textit{dependency} representations.

\begin{figure*}[t]
\centering
\includegraphics[width=0.99\textwidth]{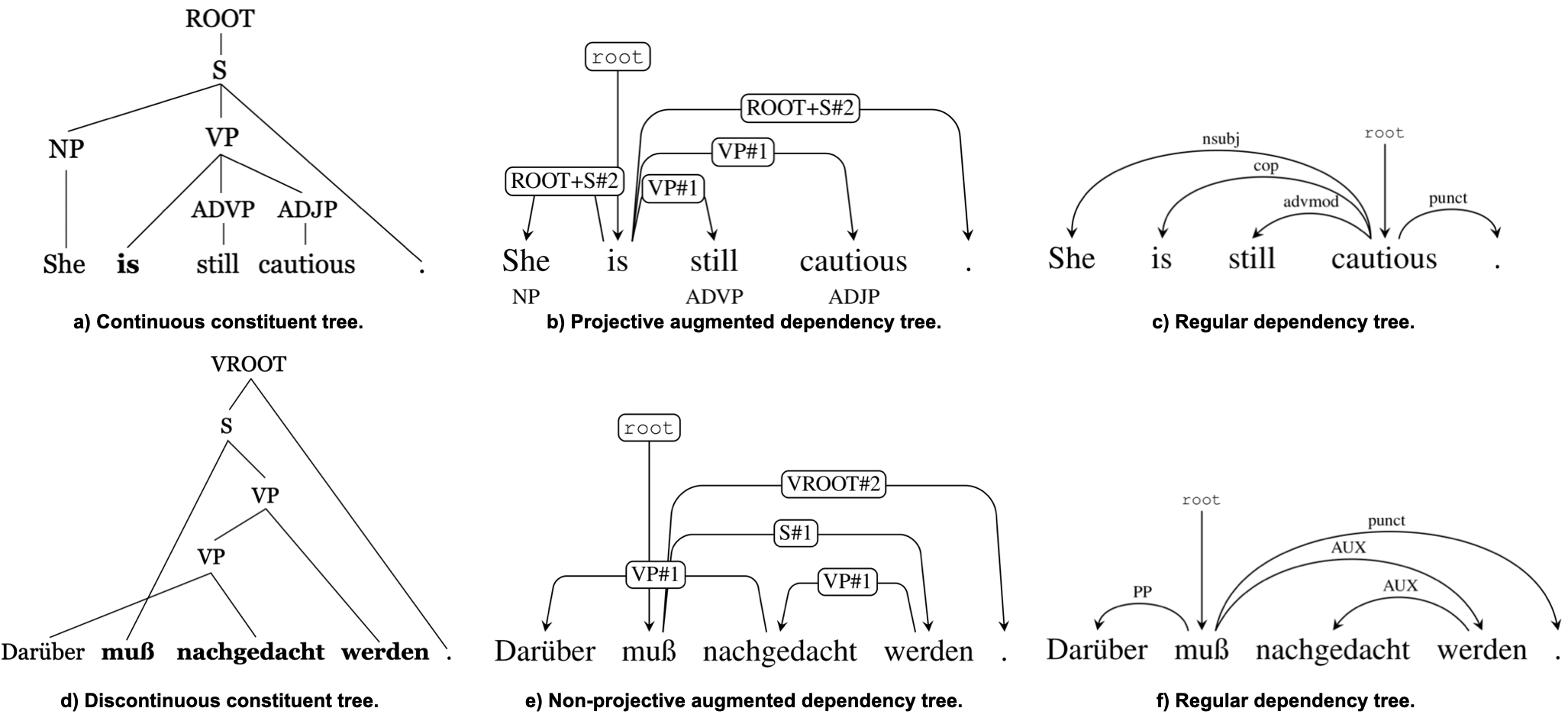}
\caption{Constituent, augmented and regular dependency representations of continuous English and discontinuous German sentences. 
Head words
of constituent trees are marked in bold. \added{Please note that regular and augmented dependency trees differs since, while head words are marked following a syntactic strategy in the augmented variant, in regular dependency trees head words are indicated according to a semantic approach.} }
\label{fig:trees}
\end{figure*}

Constituent trees, which are commonly used in tasks where span information is crucial, \new{represent the syntax of a sentence by means of constituents (also called \textit{phrases}) that hierarchically and from the bottom up group words and/or subtrees located in lower levels.}
We can find two kinds of constituent trees: \textit{continuous} and \textit{discontinuous} (described in Figure~\ref{fig:trees}(a) and (d), respectively). \new{The latter extends the former by allowing constituents with discontinuous spans, which results in phrase-structure trees with crossing branches.} These are necessary for describing some wh-movement, long-distance extractions, dislocations, cross-serial dependencies and other linguistic phenomena common in free word order languages such as German \citep{Muller2004}.

On the other hand, \new{in a dependency tree
each word of the sentence is attached to another by a directed link that describes a dependency relation between that word and its parent (also called \textit{head}). This structure} 
%composed of binary syntactic dependencies
is known for representing information closer to semantic relations and can be classified as \textit{projective} or \textit{non-projective} (depicted in Figure~\ref{fig:trees}(c) and (f), respectively).
Non-projective dependency trees 
allow crossing dependencies, and can
model the same linguistic phenomena described by discontinuous constituent trees.

Since the information described in a 
constituent tree \new{is not fully encoded into} a 
dependency tree and vice versa \citep{Kahane2015}, 
typically parsers are
exclusively trained to produce either dependency or constituent structures and, in some cases, 
restricted to the less complex continuous/projective representations.

There are a few exceptions, i.e., approaches trained to generate both constituents and dependencies. For instance, the chart parser of \citet{zhou-zhao-2019-head} generates continuous and projective structures with a single $O(n^5)$ model, and the sequence labeling parser of \citet{strzyz19} combines continuous constituents with non-projective dependency structures.\footnote{As explained in Section~\ref{sec:related}, parsers based on lexicalized grammars were also trained on both structures in the pre-deep-learning era.} In both cases, which are discussed in more detail in Section~\ref{sec:related},
representations are shown to benefit each other in terms of accuracy.

However, to our knowledge, no such joint training approaches have been defined that support both non-projective dependency trees and discontinuous constituents; and the most accurate and least computationally complex models for these formalisms are single-representation approaches: graph-based \citep{DozatM17} and transition-based \citep{Ma18,L2RPointer} models for non-projective dependencies,  and transition-based parsers \citep{coavoux2019b,coavoux2019a,DiscoPointer} for discontinuous phrase-structure trees.

In order to fill this gap,
we propose a novel multitask transition-based parser that can efficiently generate unrestricted constituent and dependency structures (i.e., discontinuous constituents and non-projective dependencies, although it can also be restricted to continuous/projective structures if desired) from a single trained model. We design an encoder-decoder neural architecture that is jointly trained across the syntactic information represented in the two formalisms by following a multitask learning strategy \citep{multitask}. Inspired by \citep{DiscoPointer}, %(hereinafter, \citep{DiscoPointer}), 
we model constituent trees as augmented dependency structures \citep{reduction} and use two separate task-specific decoders to produce both regular and augmented dependency trees. Each decoder \new{implements a Pointer Network \citep{Vinyals15} and a multi-class classifier \citep{DozatM17} to incrementally produce} labelled dependencies from left to right, as proposed by \citet{L2RPointer}.
%(hereinafter, \citep{L2RPointer}). 
Finally, the decoding runtime ($O(n^2)$) and the required memory space
of our multi-representational approach remains the same as the single-task dependency parser by \citep{L2RPointer}, since a single model is trained and the multitask learning strategy has no impact on decoding time, allowing both decoders to be run in parallel.

We test our multi-representational neural model\footnote{Source code available at \url{https://github.com/danifg/MultiPointer}.} on the continuous English and Chinese Penn Treebanks \citep{marcus93,Xue2005} and on the discontinuous NEGRA \citep{Skut1997} and TIGER \citep{brants02} datasets. In all benchmarks, our approach outperforms single-task parsers (\citep{L2RPointer}, \citep{DiscoPointer}), which proves that learning across regular dependency trees and constituent information (encoded in dependency structures) leads to gains in accuracy in both tasks, obtaining competitive results in all cases and surpassing the current state of the art
in several datasets.

The remainder of this article is organized as follows: Section~\ref{sec:con2dep} introduces the constituent-to-dependency encoding technique developed by \citet{reduction}. In Section~\ref{sec:arch}, we describe in detail the proposed multitask Pointer Network architecture. In Section~\ref{sec:experiments}, we extensively evaluate the proposed neural model on continuous/projective and discontinuous/non-projective treebanks, as well as include a thorough analysis of their performance. Section~\ref{sec:related} presents other research works that study the joint training of neural models across different syntactic formalisms. Lastly, Section~\ref{sec:conclusion} contains a final discussion.

\section{Constituent trees as dependency structures}
\label{sec:con2dep}
%\section{Constituent trees as dependencies}
Since our multitask approach is based on the dependency parser by \citep{L2RPointer}, constituent trees must be represented as dependencies in order to be processed. This was recently explored for neural discontinuous constituent parsing in \citep{DiscoPointer} by using 
the encoding by
\citet{reduction}. In this work, we extend it to continuous phrase-structure datasets, where the 
non-negligible frequency
of unary nodes requires additional processing.

\subsection{\added{Preliminaries}}

\new{Let $w_1, w_2, \dots, w_n$ be a sentence and $w_i$ the word at position $i$. A constituent tree is defined by constituents (as internal nodes) hierarchically organized over these $n$ words (as leaf nodes). Each phrase (or constituent) is defined as a tuple $(X, \mathcal{S}, w_h)$ that includes a non-terminal symbol $X$,  
the set of words $w_i$ included in its span ($\mathcal{S}$); and, $w_h$, the word in $\mathcal{S}$ that acts as head and that can be marked by using a language-specific handwritten set of rules.
For example, the head word of constituents S and VP in Figure~\ref{fig:trees}(a) is the word \textit{is}. Furthermore, we say that a constituent tree is \textit{continuous} if there are no constituents whose yield $\mathcal{S}$ is a discontinuous substring of the sentence. If this does not hold, the tree is classified as \textit{discontinuous}, and then there is at least one constituent with one or more gaps in its span. For instance, the word \textit{mu\ss} interrupts the span of constituent (VP, $\{$\textit{Dar\"uber}, \textit{nachgedacht}$\}$, \textit{nachgedacht}) in Figure~\ref{fig:trees}(d), resulting in a phrase structure with crossing branches.
Finally, constituents with exactly one child are known as \textit{unary constituents} (for instance, ROOT, NP, ADVP and ADJP in Figure~\ref{fig:trees}(a)).}

\new{Unlike constituent structures, dependency trees do not require extra internal nodes and are exclusively composed of the words $w_i$ of the sentence (plus an artificial \texttt{root} node) and binary directed links to connect them.
Each dependency link is represented as $(w_h,w_d,l)$, where $w_h$ is the head word of the dependent word $w_d$ ($h$, $d \in [1,n]$) and $l$ a dependency label. Additionally, a dependency tree is classified as \textit{projective} if we can find a directed path from $w_h$ to all words $w_i$ between words $w_h$ and $w_d$ for every dependency link $(w_h,w_d,l)$. If this does not hold, it is considered a \textit{non-projective} dependency tree, as the one with crossing arcs depicted in Figure~\ref{fig:trees}(e).}

\subsection{\added{Constituent-to-dependency conversion}}
\new{\citet{reduction} designed an encoding technique to represent a unariless constituent tree with $m$ words as a set of $m-1$ labelled dependency arcs with enriched information (plus an arc 
from
\texttt{root}),
where discontinuous phrase structures are encoded as non-projective dependency trees and continuous structures as projective trees, as shown in Figure~\ref{fig:trees}(b) and (e) for constituent trees in Figure~\ref{fig:trees}(a) and (d), respectively. To that end,
for each constituent $(X,\mathcal{S},w_h)$ with head word $w_h$, each child node $w_d$ (different from $w_h$) is encoded into the unlabelled dependency link $(w_h, w_d)$. Please note that a constituent's non-head child nodes $w_d$ might be a word or another constituent $(Y,\mathcal{G},w_d)$ with $w_d$ as head word.  
Additionally, these dependencies are augmented with an arc label that includes
the non-terminal name $X$ concatenated with an index $k$ that indicates the hierarchical order in which non-terminal nodes are built in the tree, resulting in labelled dependency arcs with the form $(w_h,w_i,X\#k)$. Index $k$ was included for those cases where several constituents share the same head word, but they are placed in the tree at a different level. For instance, constituent (S, $\{$\textit{Dar\"uber}, \textit{mu\ss}, \textit{nachgedacht}, \textit{werden}$\}$, \textit{mu\ss}) in Figure~\ref{fig:trees}(d) is represented as the augmented dependency arc (\textit{mu\ss}, \textit{werden}, S\#1) in Figure~\ref{fig:trees}(e); and constituent (VROOT, $\{$\textit{Dar\"uber}, \textit{mu\ss}, \textit{nachgedacht}, \textit{werden}, \textit{.}$\}$, \textit{mu\ss}) is encoded as (\textit{mu\ss}, \textit{.}, VROOT\#2). Both share head word \textit{mu\ss}, but the latter is built on top of the former and this must be encoded by hierarchical orders 1 and 2; otherwise, after the deconversion, the resulting structure would be a single constituent (named S or VROOT) that spans all the sentence.}

Finally, unary constituents are not directly supported by this encoding strategy. While \citet{reduction} proposed to remove all unary nodes and recover them in a post-processing step, 
 we decided to incorporate unary constituents into the resulting augmented dependency tree by collapsing non-leaf unary chains (for instance, ROOT from Figure~\ref{fig:trees}(a) into ROOT+S) and saving leaf unary nodes lost after the encoding by assigning them to words (as can be seen in Figure~\ref{fig:trees}(b) for NP, ADVP and ADJP).

\subsection{\added{Constituent trees recovery}}

\added{The original unariless constituent trees can be decoded from augmented dependency trees by, following
%starting from the root word and continuing through outgoing dependencies processing dependent words from left to right and building constituents according to the hierarchical order dictated by index $k$. 
a post order traversal, building constituents from the set of dependencies composed of each head word together with its dependents and following 
%according to 
the hierarchical order dictated by the index $k$ and non-terminal name $X$ encoded into each of the dependency labels. Due to erroneous predictions, it might be the case that heads or dependency labels are mistakenly assigned in the resulting augmented dependency tree; however, \citep{reduction}'s technique guarantees that the output is a well-formed constituent tree (which, of course, will differ from the gold tree). For instance, imagine that the word \textit{cautious} in Figure~\ref{fig:trees}(b) is erroneously attached to the word \textit{still} (instead of being connected to the verb \textit{is}), then, instead of a single flat VP with three child nodes, the resulting constituents would be two VPs (the first would have as child nodes the word \textit{is} and a second VP, which would group the words \textit{still} and \textit{cautious}). We can also find different scenarios where dependency labels are erroneously predicted, requiring ad-hoc heuristics during the recovery to deal with some inconsistencies:
\begin{itemize}
    \item \textit{Same indices, but different non-terminal names:} Note that dependency labels with the same head and at the same level (same index $k$) should share the same non-terminal name so that a flat constituent can be recovered. If this does not hold, then the dependency label of the dependent closer to the head will be the one chosen for tagging the resulting constituent. For instance, if the arcs \textit{is}$\rightarrow$\textit{still} and \textit{is}$\rightarrow$\textit{cautious} in Figure~\ref{fig:trees}(b) were tagged with labels VP\#1 and (incorrect) NP\#1, respectively; then we would use non-terminal label VP for naming the output flat constituent and NP would be discarded. Alternatively, we could consider that the non-terminal name is correct and index $k$ was wrongly predicted: in our running example, we might think that the non-terminal name of dependency label NP\#1 is correct, but the resulting constituent NP should be in a higher level (with the correct label being NP\#2, for instance). This heuristic would lead us to build a constituent NP with a nested VP. However, \citet{reduction} decided to follow a more conservative strategy that tends to produce flatter structures. 
    \item \textit{Non-nested indices in continuous parsing:} When the reverse conversion is restricted to continuous constituent trees, erroneous dependency labels might lead to discontinuous structures (even when the augmented dependency tree is projective). For example, if the arcs \textit{is}$\rightarrow$\textit{still} and \textit{is}$\rightarrow$\textit{cautious} were tagged with labels (incorrect) VP\#2 and VP\#1, respectively; then the resulting constituent would be a discontinuous constituent VP with two child nodes: the word $still$ and a non-nested VP (with a discontinuous yield composed of the words $is$ and $cautious$). This would be a well-formed phrase-structure tree in a discontinuous scenario; however, to produce continuous structures, hierarchical indices of dependent words closer to the head should always be the same or lower than adjacent and more distant siblings, thus ensuring that flat or nested continuous constituents will be obtained. In our running example, if the arcs \textit{is}$\rightarrow$\textit{still} and \textit{is}$\rightarrow$\textit{cautious} were erroneously tagged with labels VP\#2 and VP\#1, respectively; then we would decrease index 2 of dependent word $still$ until reaching the index of the adjacent and more distant sibling (the word $cautious$). In this case, the index is set to 1 and a flat constituent VP with three child nodes is built. 
\end{itemize}
} 

\added{With respect to unary recovery, it is worth noting that, while Penn treebanks present a significant amount of unary constituents, they are very uncommon in discontinuous datasets: NEGRA has no unaries at all and TIGER contains less than 1\%. Therefore, we only perform unary recovery in Penn treebanks. To do so, we simply uncollapse unary chains encoded in dependency labels and, for recovering leaf unary nodes lost after the encoding, we use a tagger in a post-processing step. More in detail, we employ the neural sequence tagger developed by \citet{yang-zhang-2018-ncrf} for assigning to each word a possible leaf unary node (or a sequence of unaries collapsed into a single tag) seen in the training dataset or the tag \texttt{NONE} (if there is no unary node above that word).}

\subsection{\added{Regular vs. augmented dependency trees}}

\added{Although both the constituent-based and regular dependency structures are
directed trees of $n$ nodes,
each 
provides exclusive information: span phrase information 
is 
included
in arc labels of the augmented variants, and regular dependency labels provide additional semantic information not described in phrase-structure trees.
Furthermore, regular dependency trees differ from augmented ones, not only in the label set, but also in the
%strategy used for marking head words during the 
conversion process. Although dependency trees are 
often
generated from constituent trees, different head-rule sets for marking head words and other transformations 
can be
applied in that process. While a set of syntactic rules are used for identifying head nodes when augmented dependency trees are produced, a semantic-based transformation is applied for choosing the semantic heads necessary for generating regular dependency structures.  This is the reason why dependency structures in Figure~\ref{fig:trees}(b) and (e) are different from Figure~\ref{fig:trees}(c) and (f), respectively: for the English example, we use the head-rule set by \citet{collins99thesis} in our constituent-to-dependency encoding, while regular dependency trees were obtained following the 
Stanford Dependencies conversion \citep{de-marneffe-manning-2008-stanford}; and, for the German sentence, the augmented dependency tree requires a non-projective stucture to fully encode the discontinuous constituent tree, while the regular dependency tree represents 
the syntax (and semantics) of 
the sentence with just a projective structure. This will 
train the 
parser across a broader variety 
of syntactic representations and notations.}

\section{Multitask Neural Architecture}
\label{sec:arch}
To develop a neural network capable of producing state-of-the-art, unrestricted constituent and dependency parses, we join two transition-based parsers recently presented under the same architecture: \citep{L2RPointer} for non-projective dependency parsing, and \citep{DiscoPointer}, an extension of the former that can produce discontinuous constituent trees. As explained before, we additionally extend the latter to also deal with continuous phrase structures and unary constituents. 

\citep{L2RPointer} relies on \textit{Pointer Networks} \citep{Vinyals15} to perform unlabelled dependency parsing.
\new{After learning the conditional probability of a sequence of numbers that represent positions from the input,
these neural networks use a mechanism of attention \citep{Bahdanau2014} to select those positions during decoding. Unlike regular sequence-to-sequence architectures, Pointer Networks do not require a fixed dictionary based on the whole training dataset, but the dictionary size is specifically defined by each input sequence length.} \new{\citet{L2RPointer} adapt Pointer Networks to implement a transition-based approach that, starting at the first word of a sentence of length $n$, sequentially attaches, from left to right, the current focus word to the pointed head word, incrementally building a well-formed dependency tree in just $n$ steps. This can be also seen as a sequence of 
$n$ SHIFT-ATTACH-$p$ transitions, each of which connects the current focus word to the head word in the pointed position $p$, and then moves the focus to the next word.
In addition, a \textit{biaffine} classifier \citep{DozatM17} jointly trained is used for predicting dependency labels. }

Inspired by \citep{L2RPointer}, 
we introduce a novel neural architecture with two \textit{task-specific decoders}:
each word of the input sentence is attached to its regular head by the first decoder, and to its augmented dependency head by the second decoder. Additionally, each decoder provides a biaffine classifier trained on its task-specific label set. Since both decoders are aligned, the resulting system requires just $n$ steps to dependency and constituent\footnote{Constituent trees are obtained after decoding resulting augmented dependency trees.} parse a sentence of length $n$, easily allowing  joint training.

More specifically, our neural architecture is composed of:
\paragraph{Shared Encoder} Each input sentence $w_1, \dots ,w_n$ is encoded by a BiLSTM-CNN architecture \citep{Ma2016}, word by word, into a sequence of \textit{encoder hidden states} $\mathbf{h}_1, \dots, \mathbf{h}_n$. In particular, a Convolutional Neural Network (CNN) is used for extracting a character-level representation of words 
($\mathbf{e}^c_i$) 
and this is concatenated with a word embedding 
($\mathbf{e}^w_i$) 
to create the vector representation $\mathbf{x}_i$ for each input word $w_i$. Additionally, POS tag embeddings 
($\mathbf{e}^p_i$) 
are used when gold POS tags are available:\footnote{As noticed by \citet{Ma18} and \citet{DiscoPointer}, the usage of predicted POS tags does not lead to gains in accuracy. Therefore, we only use POS tags in experimental settings where they are gold.} 
$$\mathbf{x}_i = \mathbf{e}^c_i \oplus \mathbf{e}^w_i \oplus \mathbf{e}^p_i$$
Then, 
the word representation 
$\mathbf{x}_i$ 
is fed one-by-one into a BiLSTM \new{for generating vector representations $\mathbf{h}_i$, which encode context information captured in both directions:}
$$ \mathbf{h}_i = \mathbf{h}_{li}\oplus\mathbf{h}_{ri} = \mathbf{BiLSTM}(\mathbf{x}_i)$$
Additionally,a  special  vector  representation $\mathbf{h}_0$, denoting the ROOT node, is prepended at the beginning of the sequence of encoder hidden states.  

\dani{Finally, we extend the encoder with deep contextualized word embeddings ($\mathbf{e}^{BERT}_i$) extracted from the pre-trained language model BERT \citep{devlin-etal-2019-bert} by directly concatenating them to the resulting basic word representation $\mathbf{x}_i$ before feeding the BiLSTM-based encoder:
$$\mathbf{x}'_i = \mathbf{x}_i \oplus \mathbf{e}^{BERT}_i;\ \mathbf{h}_i = \mathbf{BiLSTM}(\mathbf{x}'_i)$$}

\paragraph{Task-specific Decoders} Each decoder $d$ is implemented by a separate LSTM that, at each time step $t$, receives as input the encoder hidden state $\mathbf{h}_i$ of the current focus word $w_i$ and generates a \textit{decoder hidden state} $\mathbf{s}_t^d$:\footnote{Unlike \citep{L2RPointer}, we do not use other encoder hidden states as extra feature information for the decoder, since we noticed that practically the same accuracy can be achieved with this simple framework.}
$$\mathbf{s}_t^d = \mathbf{LSTM}_d(\mathbf{h}_i)$$
Additionally, a \textit{pointer layer} is implemented for each decoder by an attention vector $\mathbf{a}_t^d$ to perform unlabelled parsing.
This vector is generated by computing scores for all possible head-dependent pairs between the current focus word (represented by $\mathbf{s}_t^d$) and each word from the input (represented by encoder hidden representations $\mathbf{h}_j$ with $j \in [0,n]$). To that end, a scoring function based on the biaffine attention mechanism \citep{DozatM17} is used and, then,
 a probability distribution over the input words is computed:
$$\mathbf{v}^d_{tj} = \mathbf{score}(\mathbf{s}_t^d, \mathbf{h}_j)= f_1(\mathbf{s}_t^d)^T W f_2(\mathbf{h}_j)
+\mathbf{U}^Tf_1(\mathbf{s}_t^d) + \mathbf{V}^Tf_2(\mathbf{h}_j) + \mathbf{b};$$
$$\mathbf{a}_t^d = \mathbf{softmax}(\mathbf{v}_t^d)$$
where $W$ is the weight matrix of the bi-linear term, $\mathbf{U}$ and $\mathbf{V}$ are the weight tensors of the linear terms, $\mathbf{b}$ is the bias vector and 
$f_1(\cdot)$ and $f_2(\cdot)$ are two single-layer multilayer perceptrons (MLP) with ELU activation \citep{DozatM17}.

Each attention vector $\mathbf{a}_t^d$ will serve as a pointer to the highest-scoring position $p$ from the input, leading the parsing algorithm to create a dependency arc from the head word ($w_p$) to the current focus word ($w_i$). \added{In case this dependency arc is forbidden since it generates cycles in the already-created dependency tree,
%due to the acyclicity constraint, 
the next highest-scoring position in $\mathbf{a}_t^d$ will be considered as output instead. Furthermore, the projectivity constraint is also enforced when processing continuous treebanks, discarding arcs that produce crossing dependencies. After the decoding process (where each word is attached to another word at each step), we obtain a well-formed dependency tree where each word has a single head (except the artificial ROOT node that was not processed), with no cycles and, as a consequence of satisfying both the single-head and acyclicity constraints, all words are guaranteed to be connected.}

Finally, each decoder trains a \textit{labeler layer} (implemented as a multi-class classifier) to predict arc labels and produce a labelled dependency tree. In particular, after the pointer layer attaches the current focus word $w_i$ (represented by $\mathbf{s}_t^d$) to the pointed head word $w_p$ in position $p$ (represented by $\mathbf{h}_p$), this layer uses the same scoring function as the pointer to compute the score of each possible label for that arc and assign the highest-scoring one:
$$\mathbf{u}^{dl}_{tp} = \mathbf{score}(\mathbf{s}_t^d, \mathbf{h}_p, l)= g_1(\mathbf{s}_t^d)^T W_l g_2(\mathbf{h}_p)
+\mathbf{U}_l^Tg_1(\mathbf{s}_t^d) + \mathbf{V}_l^Tg_2(\mathbf{h}_p) + \mathbf{b}_l$$
\new{where $W_l$, $\mathbf{U}_l$, $\mathbf{V}_l$ and $\mathbf{b}_l$ are parameters distinctly used for each label $l \in \{1, 2, \dots , L\}$, being $L$ the number of labels. }
In addition, 
$g_1(\cdot)$ and $g_2(\cdot)$ are two single-layer MLPs with ELU activation.

\begin{figure}
\centering
\includegraphics[width=0.95\textwidth]{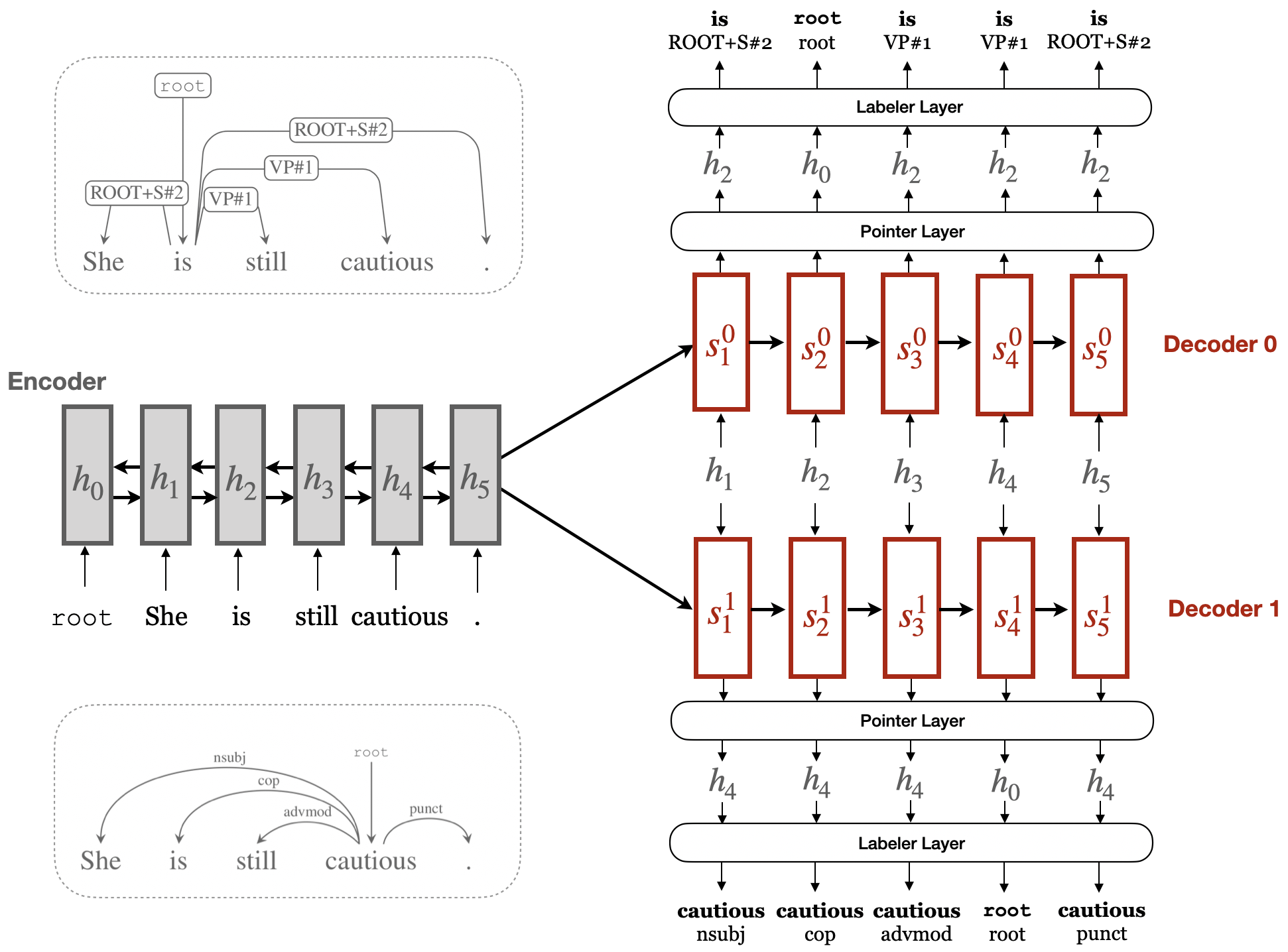}
\caption{Simplified sketch of our multitask neural architecture and 
decoding 
steps to 
parse the sentence in Figure~\ref{fig:trees}(a). {\tt Decoder 0} and {\tt Decoder 1} perform constituent-based and regular dependency parsing, respectively.}
\label{fig:network}
\end{figure}

The described transition-based algorithm can produce unrestricted non-projective dependency structures in 
$O(n^2)$
time complexity, since each decoder $d$ requires $n$ attachments to successfully parse a sentence with $n$ words, 
and at each step the attention vector $\mathbf{a}_t^d$ is computed over the whole input. Figure~\ref{fig:network} depicts
a sketch of the multitask neural architecture and the decoding procedure for parsing the sentence in Figure~\ref{fig:trees}(a).

\paragraph{Multitask Training} \new{Following a multitask learning strategy \citep{multitask}, we jointly train a single neural model for more than one task by optimizing the sum of their objectives and sharing a common encoder representation.}

\added{As both tasks use a dependency representation, the training objective of the pointer of each decoder 
is to learn the probability $P_\theta (y|x)$, where $y$ is
the correct unlabelled dependency tree for a given sentence $x$: $P_\theta (y|x)$. This probability can be factorized to the sequence of \textsc{Shift-Attach}-$p$ transitions to build $y$ (this is basically the sequence of indices $p_i$):
%a set of head-dependent pairs as follows:
$$P_\theta (y|x)=\prod_{i=1}^{n} P_\theta (p_i | p_{<i}, x)$$
%= \prod_{i=1}^{n} P_\theta (w_h | w_i,a_{<i},x)$$
%where $p_i$ denotes each arc of the dependency tree $y$ that connects each word $w_i$ to the head word $w_h$ following a left-to-right order, and 
where $p_{<i}$ represents previous predicted indices following the left-to-right order. We minimize the negative log of the probability of choosing the correct sequence of indices $p$ implemented as cross-entropy loss:
$$\mathcal{L}^d_{pointer} = -\sum_{i=1}^{n}log P_\theta (p_i,p_{<i},x)$$
%$$\mathcal{L}^d_{pointer} = -\sum_{i=1}^{n}log P_\theta (w_h | w_i,a_{<i},x)$$
Additionally, the labeler of each decoder is trained with softmax
cross-entropy to minimize the negative log likelihood of tagging with the correct label $l_i$ a given dependency arc 
defined between the head word in position $p_i$ and the dependent word in the $i^{th}$ position:
$$\mathcal{L}^d_{labeler} = -\sum_{i=1}^{n}logP_\theta (l_i|p_i, i)$$}

Then, the whole neural model is jointly trained by summing the pointer and labeler losses of each decoder:
$$\mathcal{L} = \mathcal{L}^{const}_{pointer}+\mathcal{L}^{const}_{labeler}+\mathcal{L}^{dep}_{pointer}+\mathcal{L}^{dep}_{labeler}$$

Finally, since both are considered main tasks and our goal is to train exclusively a single model, we neither use weights nor perform auxiliary-task training.

\section{Experiments}
\label{sec:experiments}
\subsection{Data}
To test our approach, we focus on parallel data, where both constituent and dependency representations are available. In particular, we conduct experiments on well-known continuous datasets: the 
English Penn Treebank (PTB) \citep{marcus93} and its Stanford Dependencies \citep{de-marneffe-manning-2008-stanford} conversion (using the Stanford parser v3.3.0)
\footnote{\url{https://nlp.stanford.edu/software/lex-parser.shtml}} 
with standard splits; 
and the Chinese Penn Treebank 5.1 \citep{Xue2005} and its converted dependency variant \citep{zhang-clark-2008-tale} with gold POS tags and two different splits: ZCTB \citep{zhang-clark-2008-tale}, for dependency parsing, and LCTB \citep{liu-zhang-2017-shift}, commonly used for constituent parsing. In addition, \new{we undertake further experiments on two broadly-used discontinuous German treebanks} and their available non-projective dependency representations: NEGRA \citep{Skut1997} with standard splits \citep{dubey2003} and TIGER \citep{brants02} with the split provided in the SPMRL14 shared task
\citep{seddah-etal-2013-overview,SPMRL}. 
For both datasets, we report results with and without gold POS tags. 

For the constituent-to-dependency encoding, \new{we identify head words on German constituents by applying the head-rule set defined by \citet{Rehbein2009} and, on English and Chinese structures, by using those developed by \citet{collins99thesis} and \citet{zhang-clark-2008-tale}, respectively.} The resulting augmented dependencies match regular variants 
by around
70\% in all languages, except for Chinese where the unlabelled augmented and regular dependency trees are exactly the same.

Following standard practice, we discard punctuation for evaluating on both Penn treebanks, using the EVALB script to report constituent accuracy. Furthermore, \new{while all tokens are considered when reporting dependency performance on German datasets, we employ discodop
\footnote{\url{https://github.com/andreasvc/disco-dop}} 
\citep{Cranenburgh2016} and ignore punctuation and root symbols
for evaluating on discontinuous constituent treebanks.}

\subsection{Settings}
Word vectors are initialized with pre-trained structured-skipgram embeddings \citep{Ling2015} for all languages and character and POS tag embeddings are randomly initialized. All of them are fine-tuned during training. POS tag embeddings are only enabled when gold information is used. 

Additionally, we report accuracy gains by augmenting our model with 
%deep contextualized word embeddings from the 
the pre-trained language model BERT \citep{devlin-etal-2019-bert}. 
\dani{Although different approaches to initialize deep contextualized word embeddings from 
%the pre-trained language model 
BERT can be found, we proceed with weights extracted from one or several layers for 
each token as a word-level representation. In addition, 
since BERT is trained on subwords,
we take the vector of each subword of an input token $w_i$ and use the average embedding as the final representation $\mathbf{e}^{BERT}_i$. In particular, we use in our experiments the pre-trained  cased German and Chinese BERT$_{\tt BASE}$ models with 12 768-dimensional hidden vectors; and uncased BERT$_{\tt LARGE}$ with 24 1024-dimensional layers for English. Depending on the specific task, some layers proved to be more beneficial than others, which is especially crucial when the resulting embeddings are not fine-tuned during training. In order to check which layers are more suitable for our tasks, we test on development sets the combination of different layers. In Table~\ref{tab:bert1}, we compare, for the English pre-trained model BERT$_{\tt LARGE}$, the accuracy obtained by averaging several groups of four consecutive layers (from last layer 24 to layer 13) and by just using weights from the second-to-last hidden layer (the simplest and commonly-used strategy, since it is less biased than the last layer to the target objectives used to train BERT). As can be seen, the combination of layers from 17 to 20 achieves the highest accuracy on both tasks and, therefore, this setup is used in our experiments on the PTB. Regarding the pre-trained models BERT$_{\tt BASE}$ for German and Chinese, we noticed that comparable accuracies can be obtained by just using weights from the second-to-last layer instead of combining the four last layers as can be seen, for instance, in Table~\ref{tab:bert2} for the NEGRA dataset. Therefore, we decided to follow the simplest configuration and use the second-to-last layer in all experiments on German and Chinese languages. We discarded other combinations such as the concatenation of several layers \new{to avoid increasing the dimension of BERT embeddings}. Finally, by adapting BERT-based embeddings to our specific tasks, our approach would certainly obtain some gains in accuracy;
however, we consider that the amount of resources necessary to that end will not justify the expensive fine-tuning of parameter-heavy BERT layers.}

\begin{table}[h]
%\begin{small}
\begin{center}
\centering
\begin{tabular}{@{\hskip 0pt}l@{\hskip 8pt}c@{\hskip 8pt}cc@{\hskip 8pt}c@{\hskip 0pt}}
\toprule
 & \multicolumn{2}{@{\hskip 0pt}c@{\hskip 0pt}}{\textbf{Regular}} & \multicolumn{2}{@{\hskip 0pt}c@{\hskip 0pt}}{\textbf{Augmented}} \\
 & \textbf{UAS} & \textbf{LAS} & \textbf{UAS} & \textbf{LAS} \\
\midrule
Layer 23  & 96.73  &  94.98 & 96.06 & 94.55 \\
Layers 21-24  & 96.69  &  94.99 & 96.03 & 94.61 \\
Layers 17-20  & \textbf{96.88}  &  \textbf{95.13} & \textbf{96.19} & \textbf{94.75} \\
Layers 13-16  & 96.71  &  94.97 & 96.08 & 94.68 \\
\bottomrule
\end{tabular}
\centering
\setlength{\abovecaptionskip}{4pt}
\caption{Accuracy comparison on regular and augmented dependency trees of the PTB development set by using weights from different BERT layers.}
\label{tab:bert1}
\end{center}
%\end{small}
\end{table}

\begin{table}[h]
%\begin{small}
\begin{center}
\centering
\begin{tabular}{@{\hskip 0pt}l@{\hskip 8pt}c@{\hskip 8pt}cc@{\hskip 8pt}c@{\hskip 0pt}}
\toprule
 & \multicolumn{2}{@{\hskip 0pt}c@{\hskip 0pt}}{\textbf{Regular}} & \multicolumn{2}{@{\hskip 0pt}c@{\hskip 0pt}}{\textbf{Augmented}} \\
 & \textbf{UAS} & \textbf{LAS} & \textbf{UAS} & \textbf{LAS} \\
\midrule
Layer 11  & \textbf{96.41}  &  95.56 & \textbf{95.04} & 94.48 \\
Layers 9-12  & 96.40  &  \textbf{95.57} & 95.02 & \textbf{94.50} \\
Layers 5-8  & 96.31  & 95.50 & 94.89 & 94.40 \\
\bottomrule
\end{tabular}
\centering
\setlength{\abovecaptionskip}{4pt}
\caption{Accuracy comparison on regular and augmented dependency trees of the NEGRA development set by using weights from different BERT layers.}
\label{tab:bert2}
\end{center}
%\end{small}
\end{table}

In each training epoch, we use the same number of examples from each task and choose the multitask model with the highest harmonic mean among Labelled Attachment Scores on augmented and regular development sets. \new{In addition, average accuracy over 3 repetitions is reported due to random initializations.}  

Finally, \new{for parameter optimization and hyper-parameter selection, we follow \citep{Ma18,DozatM17} and these are detailed in Table~\ref{tab:hyper}}. \dani{Please note that we use for the multitask variant the exact same hyper-parameters as the single-task baselines. By optimizing them to our specific multitask model, we could certainly increase performance; however, we decided to keep the same settings for a fair comparison.}  

\begin{table}[h]
\begin{footnotesize}
\centering
\begin{tabular}{@{\hskip 0pt}lc@{\hskip 0pt}}
\toprule
\textbf{Architecture hyper-parameters} & \\
\midrule
BiLSTM encoder layers & 3 \\
BiLSTM encoder size & 512 \\
LSTM decoders layers & 1 \\ 
LSTM decoders size & 512 \\
LSTM layers dropout & 0.33 \\
CNN window size & 3 \\
CNN number of filters & 50 \\
Word/POS/Character embedding dimension & 100\\
English BERT embedding dimension & 1024\\
German BERT embedding dimension & 768\\
Chinese BERT embedding dimension & 768\\
Embeddings dropout & 0.33 \\
MLP layers & 1 \\
MLP activation function & ELU \\
Arc MLP size & 512 \\ 
Label MLP size & 128 \\
UNK replacement probability & 0.5 \\
Beam size & 10 \\
%\midrule
%\textbf{Adam optimizer hyper-parameters} &\\
%\midrule
Optimizer & Adam \citep{Adam} \\
Initial learning rate & 0.001 \\
$\beta_1$, $\beta_2$ & 0.9 \\
Batch size & 32 \\
Decay rate & 0.75 \\
Gradient clipping & 5.0 \\
\bottomrule
\end{tabular}
\setlength{\abovecaptionskip}{4pt}
\caption{Model hyper-parameters.}
\label{tab:hyper}
\end{footnotesize}
\end{table}

\subsection{Results}
In Table~\ref{tab:comp}, we compare our own implementation of the single-task dependency and constituent parsers by \citet{L2RPointer} and \citet{DiscoPointer} to the proposed multitask approach. In all datasets tested, training a single model of the multi-representational parser across both syntactic representations leads to accuracy gains on both tasks.

In order to further put our approach into context, 
we also provide a comparison against 
state-of-the-art models. In Table~\ref{tab:dep}, we show how our approach outperforms 
the best 
dependency parsers
to date on the PTB and ZCTB with regular 
pre-trained 
word embeddings. Moreover, although 
some of the
included parsers 
use several parameter-heavy layers of BERT and 
additionally 
perform a task-specific 
adaptation via expensive 
fine-tuning, our approach achieves similar performance on PTB and improves over all models on ZCTB. We also outperform the single-task dependency parser by \citet{L2RPointer} with BERT,  
providing evidence that our multitask neural architecture is learning extra syntactic information that is not encoded in the pre-trained model BERT. Furthermore, Table~\ref{tab:con} shows that our novel parser obtains competitive accuracies on constituent PTB and LCTB without BERT
%non-contextualized word embeddings 
(best F-score to date on the latter), while being more efficient than $O(n^3)$ and $O(n^5)$ approaches such as \citep{kitaev-klein-2018-constituency,zhou-zhao-2019-head}. Finally, in Table~\ref{tab:disc} we show how our novel neural architecture outperforms all existing single-task parsers on the discontinuous NEGRA and TIGER datasets with regular word embeddings.

\begin{table*}
%\begin{small}
\centering
\begin{tabular}{@{\hskip 0pt}l@{\hskip 20pt}cc@{\hskip 20pt}c@{\hskip 20pt}ccc@{\hskip 0pt}}
\toprule
 & \multicolumn{2}{@{\hskip 0pt}c@{\hskip 20pt}}{\textbf{Single-Dep.}} &
 \textbf{Single-Const.} &
 \multicolumn{3}{@{\hskip 0pt}c@{\hskip 0pt}}{\textbf{Multi-Representational}} \\
\textbf{Treebank} & \textbf{UAS} & \textbf{LAS} & \textbf{F1 (LAS)} & \textbf{UAS} & \textbf{LAS} & \textbf{F1 (LAS)} \\
\midrule
PTB$_{no POS}$ & 96.06 & 94.50 & 93.29 (93.57) & \textbf{96.25} & \textbf{94.64} & \textbf{93.67 (93.93)} \\
LCTB$_{gold}$ & 93.26 & 92.67 & 88.28 (88.49) & \textbf{93.40} & \textbf{92.88} & \textbf{88.65 (88.61)} \\
ZCTB$_{gold}$ & 90.61 & 89.51 & 86.01 (84.38) & \textbf{90.79} & \textbf{89.69} & \textbf{86.09 (84.43)}  \\
NEGRA$_{gold}$ & 94.71 & 93.87 & 86.42 (92.22) & \textbf{94.80} & \textbf{94.05} & \textbf{87.30 (92.68)}\\
NEGRA$_{no POS}$ & 94.20 & 93.19 & 85.65 (91.36) & \textbf{94.33} & \textbf{93.33} & \textbf{86.78 (91.85)} \\
TIGER$_{gold}$ & 94.24 & 92.86 & 86.74 (91.81) & \textbf{94.31} & \textbf{92.90} & \textbf{87.25 (92.22)} \\
TIGER$_{no POS}$ & 93.73 & 92.27 & 85.96 (90.89) & \textbf{93.85} & \textbf{92.35} & \textbf{86.61 (91.36)} \\
\bottomrule
\end{tabular}
\centering
\setlength{\abovecaptionskip}{4pt}
\caption{Accuracy comparison of single-task baseline parsers to the proposed multi-representational approach in both constituent and dependency parsing. We report Labeled Attachment Scores (LAS) and Unlabeled Attachment Scores (UAS) for dependency parsing and, for constituent parsing, the LAS on the augmented dependency trees and F-score on the post-decoding constituent structure. The corresponding standard deviations over 3 runs for each score are reported in Table~\ref{tab:deviations}.}
\label{tab:comp}
%\end{small}
\end{table*}

\begin{table*}
%\begin{small}
\centering
\begin{tabular}{@{\hskip 0pt}l@{\hskip 20pt}cc@{\hskip 20pt}c@{\hskip 20pt}ccc@{\hskip 0pt}}
\toprule
 & \multicolumn{2}{@{\hskip 0pt}c@{\hskip 20pt}}{\textbf{Single-Dep.}} &
 \textbf{Single-Const.} &
 \multicolumn{3}{@{\hskip 0pt}c@{\hskip 0pt}}{\textbf{Multi-Representational}} \\
\textbf{Treebank} & \textbf{UAS} & \textbf{LAS} & \textbf{F1 (LAS)} & \textbf{UAS} & \textbf{LAS} & \textbf{F1 (LAS)} \\
\midrule
PTB$_{no POS}$ & $\pm$0.03 & $\pm$0.04 & $\pm$0.06 ($\pm$0.04) & $\pm$0.04 & $\pm$0.04 & $\pm$0.05 ($\pm$0.03) \\
LCTB$_{gold}$ & $\pm$0.08 & $\pm$0.09 & $\pm$0.06 ($\pm$0.04) & $\pm$0.07 & $\pm$0.08 & $\pm$0.09 ($\pm$0.07) \\
ZCTB$_{gold}$ & $\pm$0.07 & $\pm$0.05 & $\pm$0.07 ($\pm$0.06) & $\pm$0.08 & $\pm$0.06 & $\pm$0.07 ($\pm$0.05) \\
NEGRA$_{gold}$ & $\pm$0.03 & $\pm$0.06 & $\pm$0.06 ($\pm$0.04) & $\pm$0.02 & $\pm$0.03 & $\pm$0.04 ($\pm$0.02) \\
NEGRA$_{no POS}$ & $\pm$0.04 & $\pm$0.04 & $\pm$0.05 ($\pm$0.03) & $\pm$0.06 & $\pm$0.04 & $\pm$0.06 ($\pm$0.03) \\
TIGER$_{gold}$ & $\pm$0.04 & $\pm$0.05 & $\pm$0.06 ($\pm$0.04) & $\pm$0.03 & $\pm$0.05 & $\pm$0.04 ($\pm$0.02) \\
TIGER$_{no POS}$ & $\pm$0.07 & $\pm$0.05 & $\pm$0.06 ($\pm$0.06) & $\pm$0.05 & $\pm$0.04 & $\pm$0.07 ($\pm$0.05) \\
\bottomrule
\end{tabular}
\centering
\setlength{\abovecaptionskip}{4pt}
\caption{Standard deviations of scores in Table~\ref{tab:comp} over 3 runs.}
\label{tab:deviations}
%\end{small}
\end{table*}

\begin{table}[tbp]
%\begin{small}
\begin{center}
\centering
\begin{tabular}{@{\hskip 0pt}l@{\hskip 8pt}c@{\hskip 8pt}cc@{\hskip 8pt}c@{\hskip 0pt}}
\toprule
 & \multicolumn{2}{@{\hskip 0pt}c@{\hskip 0pt}}{\textbf{PTB}} & \multicolumn{2}{@{\hskip 0pt}c@{\hskip 0pt}}{\textbf{ZCTB}} \\
\textbf{Parser} & \textbf{UAS} & \textbf{LAS} & \textbf{UAS} & \textbf{LAS} \\
\midrule
\citet{Wang2016} & 94.08 & 91.82 & 87.55 & 86.23\\
\citet{Cheng2016} & 94.10 & 91.49 & 88.1\hphantom{0} & 85.7\hphantom{0} \\
\citet{Kuncoro2016} & 94.26 & 92.06 & 88.87 & 87.30 \\
\citet{zhang-etal-2016-probabilistic} & 93.42 & 91.29 & 87.65 & 86.17 \\
\citet{Zhang17} & 94.10 & 91.90 & 87.84 & 86.15 \\
\citet{Ma2017} & 94.88 & 92.96 & 89.05 & 87.74\\
\citet{DozatM17} & 95.74 & 94.08 & 89.30 & 88.23 \\
\citet{li-etal-2018-seq2seq} & 94.11 & 92.08 & 88.78 & 86.23 \\
\citet{Ma18} & 95.87 & 94.19 & 90.59 & 89.29 \\
\citet{ji-etal-2019-graph} & 95.97 & 94.31 & - & - \\
\citet{L2RPointer} &  96.04 &  94.43 & - & -  \\
\citet{zhou-zhao-2019-head} & 96.09 & \textbf{94.68} & - & - \\
\citet{li2019global} & 95.83 & 94.54 & 90.47 & 89.44 \\
\citet{zhang-etal-2020-efficient} & 96.14 & 94.49 & - & - \\
\textbf{This work} & \textbf{96.25} & 94.64 & \textbf{90.79} & \textbf{89.69} \\
\hdashline[1pt/1pt]
\textit{+BERT} \\
\ \ \ \citet{L2RPointer} &  96.91 &  95.35 & 92.58 & 91.42  \\
\ \ \ \citet{li2019global} & 96.44 & 94.63 & 90.89 & 89.73 \\
\ \ \ \citet{li2019global}$^*$ & 96.57 & 95.05 & - & - \\
\ \ \ \citet{zhou-zhao-2019-head}$^*$ & \textbf{97.00} & 95.43 & 91.21 & 89.15 \\
\ \ \ \textbf{This work} & 96.97 & \textbf{95.46} & \textbf{92.78} & \textbf{91.65} \\
\bottomrule
\end{tabular}
\centering
\setlength{\abovecaptionskip}{4pt}
\caption{Accuracy comparison of state-of-the-art dependency parsers on PTB and ZCTB. Models that fine-tune BERT are marked with $^*$. \dani{Since in the original work \citep{L2RPointer} performance with BERT was not reported, we run our own implementation of the single-task dependency parser enhanced with BERT-based embeddings and include it in the second block as ``\citet{L2RPointer}''}.} 
\label{tab:dep}
\end{center}
%\end{small}
\end{table}

\begin{table}[tbp]
\begin{center}
\centering
\begin{tabular}{@{\hskip 0pt}lcc@{\hskip 0pt}}
\toprule
\textbf{Parser} & \textbf{PTB} & \textbf{LCTB}\\
\midrule
\citet{dyer-etal-2016-recurrent}  & 91.2\hphantom{0} & 84.6\hphantom{0}\\
\citet{cross-huang-2016-span} & 91.3\hphantom{0} & -\\
\citet{liu-zhang-2017-shift} & 91.7\hphantom{0} & 85.5\hphantom{0} \\
\citet{liu-zhang-2017-order} & 91.8\hphantom{0} &  86.1\hphantom{0}\\
\citet{fernandez-gonzalez-gomez-rodriguez-2018-dynamic} & 92.0\hphantom{0} & 86.6\hphantom{0} \\
\citet{stern-etal-2017-minimal} & 91.8\hphantom{0} & -\\
\citet{stern-etal-2017-effective} & 92.56 & -\\
\citet{shen-etal-2018-straight} & - & 86.5\hphantom{0} \\ 
\citet{fried-klein-2018-policy} & 92.2\hphantom{0} & 87.0\hphantom{0} \\
\citet{gaddy-etal-2018-whats} & 92.08 & - \\
\citet{teng-zhang-2018-two} & 92.4\hphantom{0} & 87.3\hphantom{0} \\
\citet{kitaev-klein-2018-constituency} & 93.55 & -\\
\citet{zhou-zhao-2019-head} & \textbf{93.78} & - \\
\textbf{This work} & 93.67 & \textbf{88.65} \\
%\hline
\hdashline[1pt/1pt]
\textit{+BERT} \\
\ \ \ \citet{kitaev-etal-2019-multilingual}$^*$ & 95.59 & 91.75 \\
\ \ \ \citet{zhou-zhao-2019-head}$^*$ & \textbf{95.84} & \textbf{92.18} \\
\ \ \ \textbf{This work} & 95.23 & 90.20 \\
\bottomrule
\end{tabular}
\centering
\setlength{\abovecaptionskip}{4pt}
\caption{F-score comparison of state-of-the-art constituent parsers on PTB and LCTB. Models that fine-tune BERT are marked with $^*$.}
\label{tab:con}
\end{center}
\end{table}

\subsection{Analysis}
In order to obtain insight into why the multi-representational variant is outperforming single-task parsers in both tasks,\footnote{Apart from the widely-proven benefits of using multitask learning as a regularization method to avoid overfitting.} we conduct an error analysis relative to structural factors.

\begin{figure*}[h]
\centering
\includegraphics[width=\textwidth]{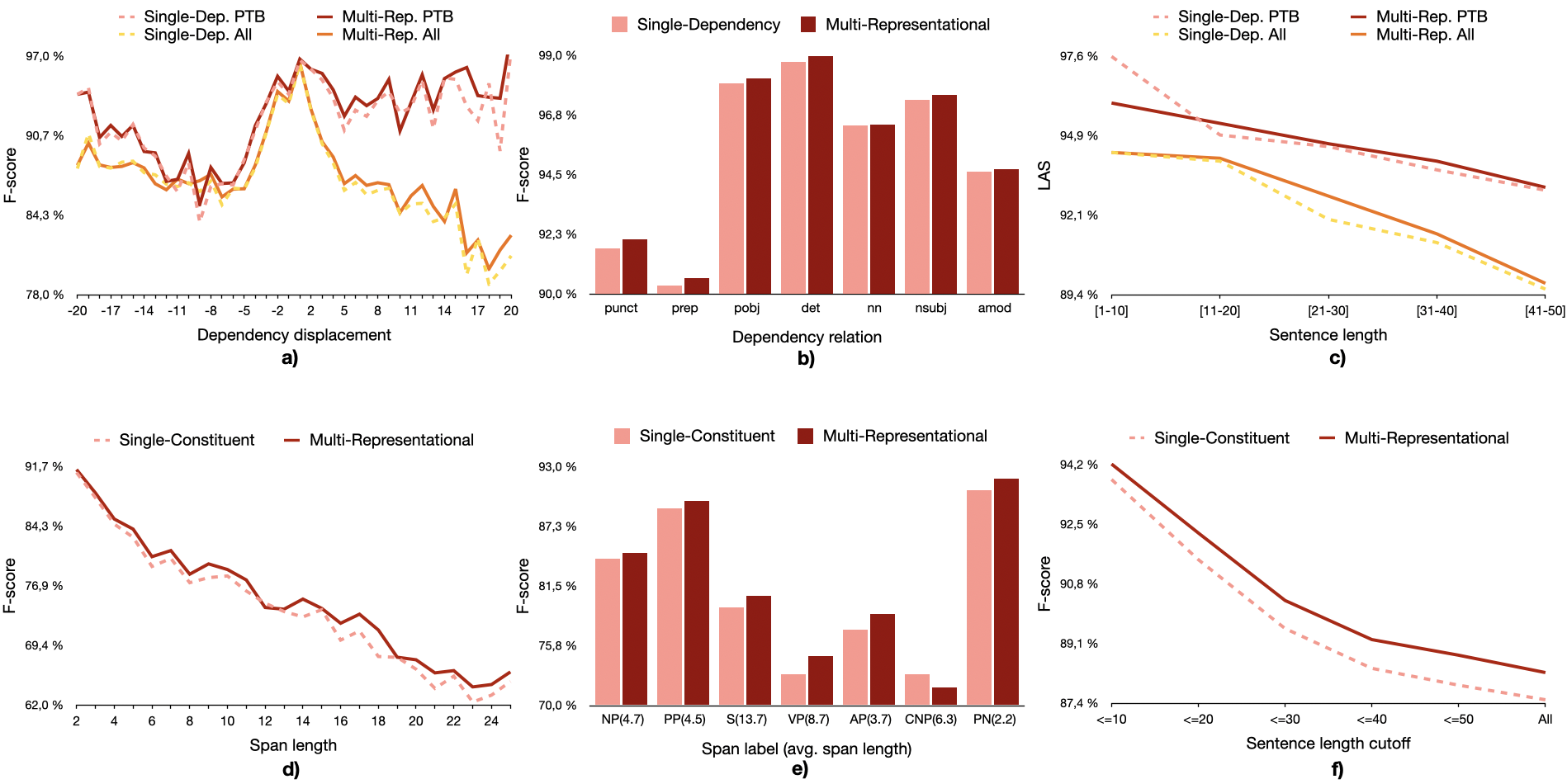}
\caption{Parsing performance of the single-task and the multi-representational parsers relative to length and structural factors.}
\label{fig:analysis}
\end{figure*}

For the dependency parsing task, we show in Figure~\ref{fig:analysis}(a) the F-score relative to dependency displacements (i.e., signed distances) on the PTB and on the concatenation of all datasets,\footnote{We discard German datasets with gold PoS tags.} Figure~\ref{fig:analysis}(b) reports the performance on common dependency relations on PTB and Figure~\ref{fig:analysis}(c) shows the accuracy of both approaches relative to sentence lengths on PTB and on all datasets together. From these results, we can point out that the multitask parser is performing 
better on longer leftward dependency arcs (with positive displacement) and on longer sentences,
improving over the single-task system in all frequent dependency relations.

Regarding constituent parsing, we specifically analyze performance on both discontinuous German datasets together, where the multi-representational model significantly outperforms the single-task approach. Firstly, we report in Table~\ref{tab:disc} \new{an F-score exclusively measured on discontinuous constituents (DF1)}, showing a notable performance on discontinuous structures (probably thanks to the joint training with regular non-projective dependency structures).
Additionally, Figure~\ref{fig:analysis}(d) plots the F-score on span identification for different lengths, Figure~\ref{fig:analysis}(e) shows the performance by span labels and Figure~\ref{fig:analysis}(f) measures the accuracy
of both approaches on different sentence length cutoffs.
It can be noticed that the multitask variant achieves higher performance when spans are larger and sentences tend to be longer, being only less accurate than the single-task parser on Coordinated Noun Phrases (CNP), where, in this particular case, a disagreement in notation between constituent and dependency representations\footnote{In the regular dependency version, a CNP structure is represented by attaching the second noun to the conjunction and the latter to the first noun, while in the augmented variant, the first noun and the conjunction are both attached to the second noun.} might be misleading the multitask approach. 

All this provides some evidences that learning across syntactic representations is tackling the main weakness of the transition-based sequential decoding:
the impact of error propagation on the performance on large constituents and long sentences. Moreover, the information exclusively encoded by each formalism (span phrase information in constituent trees and semantic relations in dependency structures) may complete each other and provide an additional guidance not only in final decoding steps (where the parser is more prone to make a mistake due to error propagation), but also in creating those structures that are less frequent in some of the two representations (as happens with long leftward dependency arcs in languages such as English).

It is also worth mentioning that even on Chinese datasets (where augmented and regular dependencies are the same) our approach benefits from learning across both structures, meaning that both constituent-based and regular dependency label sets provide useful syntactic information.

\begin{table}[tbp]
\centering
\begin{tabular}{@{\hskip 0pt}l@{\hskip 3pt}c@{\hskip 5pt}c@{\hskip 10pt}c@{\hskip 5pt}c@{\hskip 0pt}}
\toprule
& \multicolumn{2}{c}{\textbf{NEGRA}}
& \multicolumn{2}{c}{\textbf{TIGER}}
\\
\textbf{Parser} & \textbf{F1} & \textbf{DF1} & \textbf{F1} & \textbf{DF1} \\
\midrule
\scriptsize{\textit{(Predicted/Without PoS tags)}}\\
\citet{reduction}   & 77.0 & - &  77.3 & - \\
\citet{Versley2016}     & - & - &  79.5 & - \\
\citet{stanojevic2017}     & - & - &   77.0 & - \\
\citet{coavoux2017}     & - & - &   79.3 & - \\
\citet{coavoux2019a}     & 83.2 & 54.6 &   82.7 & 55.9 \\
\citet{coavoux2019b}     & 83.2 & 56.3 &   82.5 & 55.9 \\
\citet{stanojevic-steedman-2020-span} & 83.6 & 50.7 & 83.4 & 53.5  \\
\citet{vilares-gomez-rodriguez-2020-discontinuous} & 75.6 & 34.6 & 77.5 & 39.5 \\
\citet{DiscoPointer} & 85.7 & 58.6 & 85.7 & 60.4 \\
\citet{Corro2020SpanbasedDC} & 86.3 & 56.1 & 85.2 & 51.2 \\
\textbf{This work} & \textbf{86.8} & \textbf{69.5}  & \textbf{86.6} & \textbf{62.6} \\
\hdashline[1pt/1pt]
\textit{+BERT} \\
\ \ \ \citet{vilares-gomez-rodriguez-2020-discontinuous}$^*$ & 83.9 & 45.6 & 84.6 & 51.1 \\
\ \ \ \citet{Corro2020SpanbasedDC}$^*$ & \textbf{91.6} & 66.1 & \textbf{90.0} & 62.1 \\
\ \ \ \citet{fernandezgonzalez2021reducing}$^*$ & 90.4 & 66.5 & 88.5 & 62.7 \\
\ \ \ \textbf{This work} & 91.0 & \textbf{76.6} & 89.8 & \textbf{71.0} \\
\midrule
\scriptsize{\textit{(Gold PoS tags)}}\\ 
\citet{maier2015} & 77.0 & 19.8 & 74.7 & 18.8 \\
\citet{reduction}     & 80.5 & - & 80.6 & - \\
\citet{maier2016}    & - & - & 76.5 & - \\
\citet{corro2017}  & - & - & 81.6 & - \\
\citet{stanojevic2017}     & 82.9 & - & 81.6 & - \\
\citet{coavoux2017}     & 82.2 & 50.0 & 81.6 & 49.2 \\
\citet{gebhardt2018}     & - & - &   75.1 & - \\
\citet{morbitz2020supertaggingbased}  & 82.8 & 52.9 & 81.8 & 54.6 \\
\citet{vilares-gomez-rodriguez-2020-discontinuous} & 77.1 & 36.5 & 79.2 & 40.1 \\
\citet{DiscoPointer} & 86.1 & 59.9 & 86.3 & 60.7 \\
\textbf{This work} & \textbf{87.3} & \textbf{71.0} & \textbf{87.3} & \textbf{64.2} \\
\bottomrule
\end{tabular}
\centering
\setlength{\abovecaptionskip}{4pt}
\caption{F-score and Discontinuous F-score (DF1) comparison of state-of-the-art discontinuous constituent parsers on NEGRA and TIGER. Models that fine-tune BERT are marked with $^*$.
}
\label{tab:disc}
\end{table}

\added{Finally, the multitask approach achieves lower accuracies on continuous constituent datasets 
since the encoding technique by \citet{reduction} cannot directly handle unary nodes (which are collapsed
%, increasing the label set size, 
or, in case of leaf unary nodes, assigned with a regular sequence tagger), losing some accuracy in continuous treebanks where the amount of this kind of structures is significant: 19.69\% and 19.09\% of the constituents on the PTB training and development sets, respectively, are unary nodes. One consequence of encoding unaries by collapsing them is that, while the labeler on regular dependency trees deals with 47 different dependency labels on the PTB, the labeler on augmented dependency structures manages 188 different tags (104 of them being generated for encoding 
%collapsed non-leaf 
unary nodes). 
On the contrary, in discontinuous datasets such as TIGER (where unary nodes are discarded due to their low frequency), the regular label set size is 45 and the augmented version has 83. This significant increase on augmented dictionary sizes for processing continuous datasets might penalize the labeler's performance and affect final accuracy, especially
%: 98.39\% for the regular labeler against 98.05\% for the augmented labeler on the test set. This might seem a narrow difference; however, it is worth mentioning that, in the used 
in an encoding technique where dependency labels have a crucial role during constituent recovery.
%and that accuracies on PennTreebanks are already remarkably high and this loss in label accuracy might make the difference. 
Additionally, the recovery of leaf unary nodes (the 73.55\% of total unaries from PTB development set for example) lost after the constituent-to-dependency conversion  has a greater impact on final accuracy. The tagger in charge of that has to face a complex task, since the amount of words with unary constituents on top is scarce on the training set (88.85\% of words are tagged with \texttt{NONE} and, since a sequence of leaf unaries is collapsed into a single tag as done for non-leaf unary nodes, the model has to deal with a large dictionary size of 54 tags), hindering the adequate training of the tagger. While it achieves a good overall accuracy (for instance, 98.65\% on the PTB development set), a worse performance is obtained when only considering words with attached unary nodes (just the 10.59\% of total words): 92.56\% recall, 91.82\% precision and 92.19\% F-score on the PTB development set. It might seem that this performance is good enough; however, it means that tagging errors are more than 5 times as frequent in words associated with unary nodes compared to the overall error rate, and its impact on the final parsing accuracy is significant taking into account that scores on Penn treebanks are remarkably high. Despite all that, our approach obtains the best accuracy to date among all existing transition-based parsers in 
both continuous and discontinuous 
constituent structures, and it is on par with state-of-the-art models such as \citep{kitaev-klein-2018-constituency} and
\citep{zhou-zhao-2019-head}.}

\section{Related work}
\label{sec:related}
It is known that parsers based on lexicalized grammar  are trained using both constituent and unlabeled dependency information. This includes classic chart parsers \citep{collins-2003-head} as well as lexicalized parsers that build dependencies with reduce transitions, such as \citep{Crabbe2015}, which can generate both structures. These are restricted to dependencies that are directly inferred from the lexicalized constituent trees. In this sense, the multitask approach is more flexible, as it does not have that limitation and one can use dependencies and constituents from different sources.

In the deep learning era, there have been a few recent attempts to jointly train a neural model across constituent and dependency trees, producing, during decoding, both syntactic representations from a single model.

In particular, \citet{strzyz19} propose a multitask sequence labelling architecture that, by representing constituent and dependency trees as linearizations \citep{gomez-rodriguez-vilares-2018-constituent,strzyz-etal-2019-viable}, can 
learn and 
perform parsing in both formalisms as joint tasks. While being a linear 
and fast 
parser, the 
parsing 
accuracy provided by this approach is notably behind the state of the art (even training separate models by performing an auxiliary-task learning for each formalism) and the linearization strategy used for constituent parsing is restricted to continuous structures.

\citet{zhou-zhao-2019-head} also explore the benefits of training a model across syntactic representations. They propose to integrate dependency and constituent information into a simplified variant of the Head-Driven Phrase Structure Grammar formalism (HPSG). Then, to implement a HPSG parser, they modify the constituent chart-based parser by \citep{kitaev-klein-2018-constituency} that employs an $O(n^5)$ CKY-style algorithm \citep{stern-etal-2017-effective} for decoding.\footnote{They also propose a $O(n
^3)$ decoding method that achieves worse accuracy.} Although their approach can produce both syntactic structures at the same time and achieve state-of-the-art accuracies on PTB and CTB treebanks,  their parser is bounded to produce continuous and projective structures with a high runtime complexity.

Our approach can handle any kind of constituent and dependency structures and provides an efficient runtime complexity, crucial for some downstream applications.

\section{Conclusions and Future Work}
%\section{Conclusions}
\label{sec:conclusion}
We propose a novel encoder-decoder neural architecture based on Pointer Networks that, after being jointly trained on regular and constituent-based dependency trees, can syntactically parse a sentence to 
both constituent and dependency trees.
Apart from just requiring to train a single model, our approach can produce not only the simplest continuous/projective trees, but also discontinuous/non-projective structures in just $O(n^2)$ runtime. We test our parser on the main dependency and constituent benchmarks, obtaining competitive results in all cases and reporting state-of-the-art accuracies in several datasets.

As future work, we plan to perform auxiliary-task learning and train a separate model for each task, testing different weights for the loss computation. This will lose the advantage of training a single model to undertake both tasks, but will certainly lead to further improvements in accuracy.

\section*{Acknowledgments}
We acknowledge the European Research Council (ERC), which has funded this research under the European Union’s Horizon 2020 research and innovation programme (FASTPARSE, grant agreement No 714150), ERDF/MICINN-AEI (ANSWER-ASAP, TIN2017-85160-C2-1-R; SCANNER-UDC, PID2020-113230RB-C21), Xunta de Galicia (ED431C 2020/11), and Centro de Investigaci\'on de Galicia ``CITIC'', funded by Xunta de Galicia and the European Union (ERDF - Galicia 2014-2020 Program), by grant ED431G 2019/01. Funding for open access charge: Universidade da Coruña/CISUG.

% To print the credit authorship contribution details
\printcredits

%% Loading bibliography style file
%\bibliographystyle{model1-num-names}
\bibliographystyle{cas-model2-names}

% Loading bibliography database
\bibliography{anthology,main,bibliography}

\end{document}